# Floating Displacement−Force Conversion Mechanism as a Robotic Mechanism

Kenjiro TADAKUMA*, Tori SHIMIZU, Sosuke HAYASHI, Eri TAKANE,

Masahiro WATANABE, Masashi KONYO, and Satoshi TADOKORO

*Abstract*— To attach and detach permanent magnets with an operation force smaller than their attractive force, Internally-Balanced Magnetic Unit (IB Magnet) has been developed. The unit utilizes a nonlinear spring with an inverse characteristic of magnetic attraction to produce a balancing force for canceling the internal force applied on the magnet. This paper extends the concept of shifting the equilibrium point of a system with a small operation force to linear systems such as conventional springs. Aligning a linear system and its inverse characteristic spring in series enables a mechanism to convert displacement into force generated by a spring with theoretically zero operation force.

To verify the proposed principle, the authors realized a prototype model of inverse characteristic linear spring with an uncircular pulley. Experiments showed that the generating force of a linear spring can be controlled by a small and steady operation force.

## I. Introduction

There were several researches and development of the converter mechanisms its convert torque to torque, motion to motion and so on(e.g. [1]-[3]). On the other hand, conventionally, an adsorption mechanism using a magnet called internally balanced magnetic unit that can perform adsorption and desorption with a small operating force when adsorbing to a ferromagnetic material such as steel has been proposed [4]. The principle diagram is shown in Fig. The principle of this mechanism is to accumulate the energy generated at the time of adsorption in the spring and use it for the energy at the time of removal. Since the attraction force changes nonlinearly with respect to the displacement, it is necessary to use a nonlinear spring having the opposite characteristic in order to recover the energy. Fig. 4 (a) shows the repulsive force Inverse Fm of the nonlinear spring formed by the combination of the attractive force Fm of the magnet and the leaf spring. As a result, since the forces are balanced at an arbitrary displacement, the operating force is zero.

In this research, we propose a floating displacement / force conversion mechanism, not only the displacement and force characteristics of the magnet but also the general elastic element can be expanded. As shown in Fig. 2, arranging the elastic element g (x) and its inverse characteristic element-g (x) in series and providing a balance point can generate displacement of the elastic element with a small operating force it can. This makes it possible to apply to mechanical elements other than magnet attracting force, recovering the energy necessary for compressing and tensioning the element, if it is theoretically not the energy loss, it is possible to convert the displacement purely to force Become.

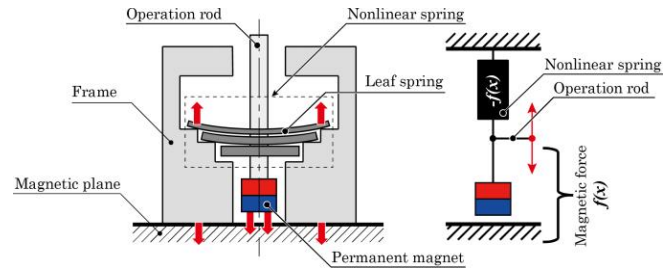

Fig. 1 Internally-Balanced Magnetic Unit.

## 2. Floating Displacement – Force Conversion Mechanism

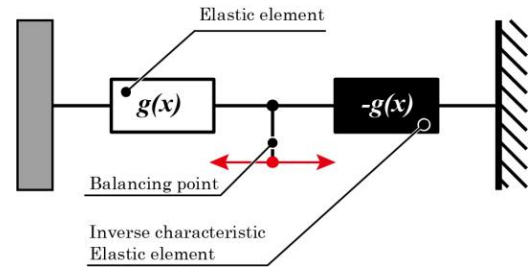

Figure 1. Force compensation of elastic element.

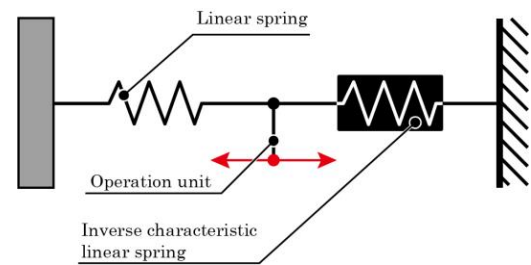

Figure 2. Principles of float converter.

* This work was supported in part by Japan Science and Technology Agency "Impulsing Paradigm Change Through Disruptive Technologies Program: Tough Robotics Challenge." (Corresponding author: Kenjiro Tadakuma.) The authors are with the Graduate School of Information Sciences, Tohoku University, Sendai 980-8579, Japan (e-mail:;

tadakuma@rm.is.tohoku.ac.jp; shimizu.tori@rm.is.tohoku.ac.jp; hayashi.sosuke@rm.is.tohoku.ac.jp;
watanabe.masahiro@rm.is.tohoku.ac.jp; konyo@rm.is.tohoku.ac.jp; tadokoro@rm.is.tohoku.ac.jp)

$$f(x) + g(x) = 0 \quad (1)$$

$$g(x) = -f(x) \quad (2)$$

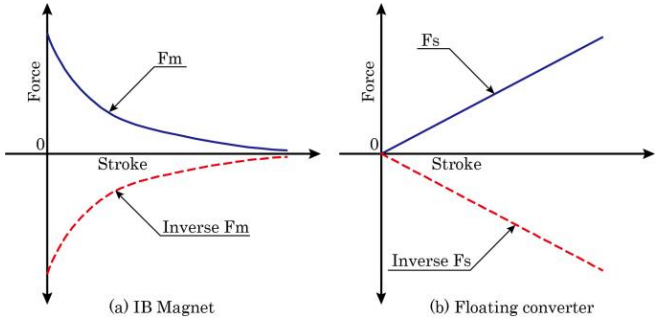

Figure 3. The relation between unit and invert unit force.

## A. Basic Principle

Fig. 3 shows a conceptual diagram of a floating displacement / force converter. Apply a linear spring and operate the spring with a very small operating force. It consists of the linear spring on the left and the inverse characteristic linear spring on the right, and manipulates the connecting part of the center which is the balance point of the force. With this configuration, the displacement of the operating part can be taken out as force, and the relationship is linear. As shown in Fig. 4 (b), since the linear spring and the element having the inverse characteristic are arranged in series, it can be seen that when the top and bottom are added up by each displacement of the spring, it becomes 0. In other words, energy is exchanged between the left and right systems.

## B. Inverse Spring Configuration Example

In order to realize the floating displacement / force conversion mechanism, it is necessary to realize the inverse characteristic of the linear spring. As methods, 1) a method using a non-circular pulley and a spring or a weight, 2) a method of adjusting the width and rigidity of the constant load spring for each displacement, and 3) a method using a cam or a link. Since non-circular pulleys can set arbitrary angle and radius, they are used for self-weight compensation [5] of robot arm and constant load spring [6] for muscle training machine. In addition, as a structure close to the inverse characteristic spring, compound bow has been devised in which the operating force becomes minute when the string is pulled [7]. In this research, we adopted a system using a non-circular pulley and a weight in consideration of the ease of prototyping in the configuration of the reverse spring.

## C. Application Example: Robot Gripper

A robot gripper as shown in Fig. 5 is conceivable as an application example making full use of the features of the floating displacement / force converter. In grasping a rigid body, it is possible to realize high speed / high gripping force, and it can be predicted that consumption of energy during grasping and gripping force control has a minute characteristic. The gripper has a configuration in which a displacement / force converter that controls force and a high-speed positioning mechanism are arranged in series to the grasped object (e.g. [8]). High lead positioning leads Long feed screws have back drivability, so reverse rotation occurs when the gripping force increases. Therefore, reverse torque should be released to the frame by the torque diode. Alternatively, by developing a mechanism having the same components as the torque diode in the direct acting part, it can be applied to mechanisms other than the feed screw.

As a gripping method, (1) the positioning mechanism operates until x is slightly opened, and (2) the displacement / force converter grips the object with an arbitrary holding force. Also, when the distance x between the gripper and the converter $x \neq 0$, the equilibrium point shifts and it is thought that force is required for the operation of the converter. In actual machine experiments to be described later, x was given as displacement, and the force required for the operation was measured.

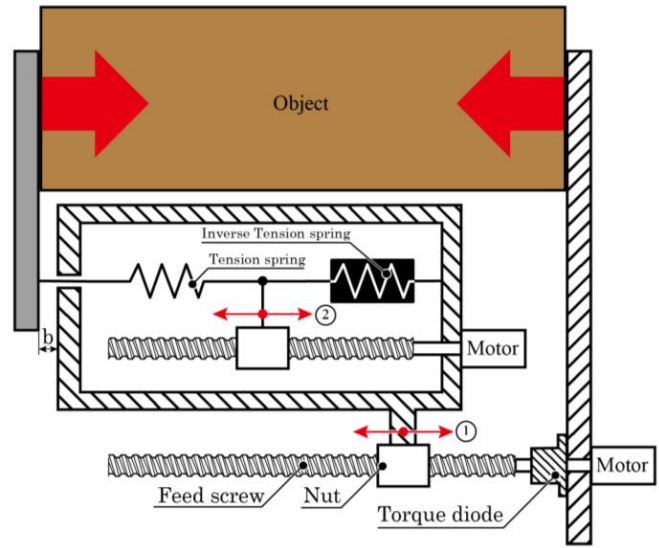

Fig. 5. Application to robot hand.

## 3. IMPLEMENTATION OF REAL MACHINE

We implemented real devices of floating displacement / force converter we invented. A tension spring is applied to the linear spring. Figure 6 shows the structure. In this mechanism, the load mg generated by the weight is converted to the inverse characteristic of the tension spring by the non-circular pulley. In addition, the non-circular pulley is connected to the circular pulley, and converts the rotation angle into the displacement of the linear spring.

In order to realize the inverse characteristic of the tension spring, the non-circular pulley was designed to have a shape in which the angle θ and the radius r are proportional. When θ is in the range of 0 ° - 345 °, r is 10 - 40 mm.

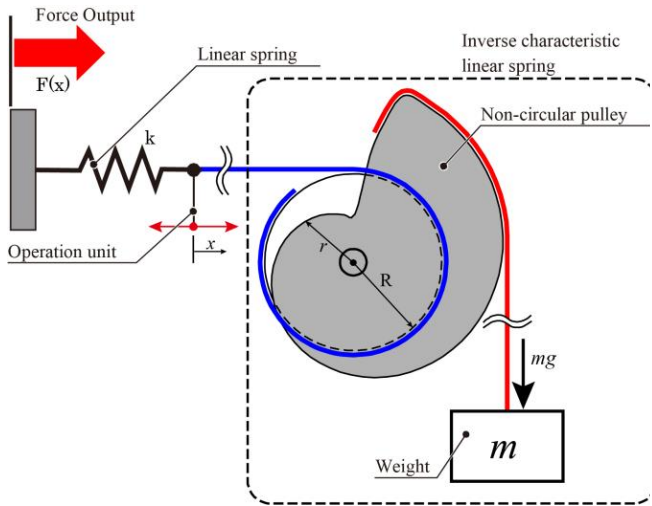

Fig. 6. Configuration example of inverse spring

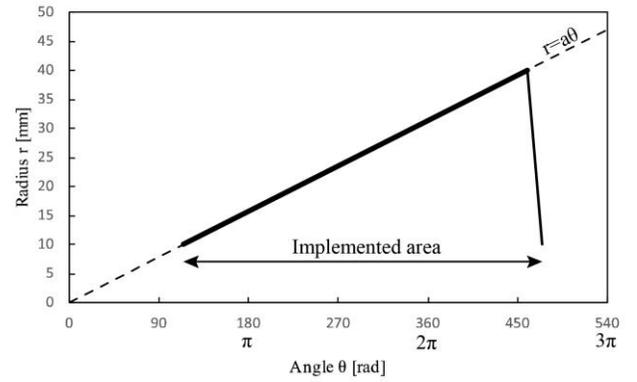

Fig. 8. The inclination of the non-circular pulley and the area produced.

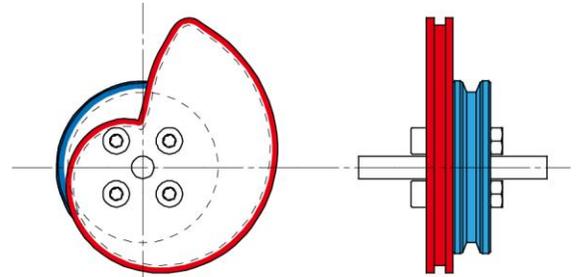

Fig. 9. Specific composition of the pulley.

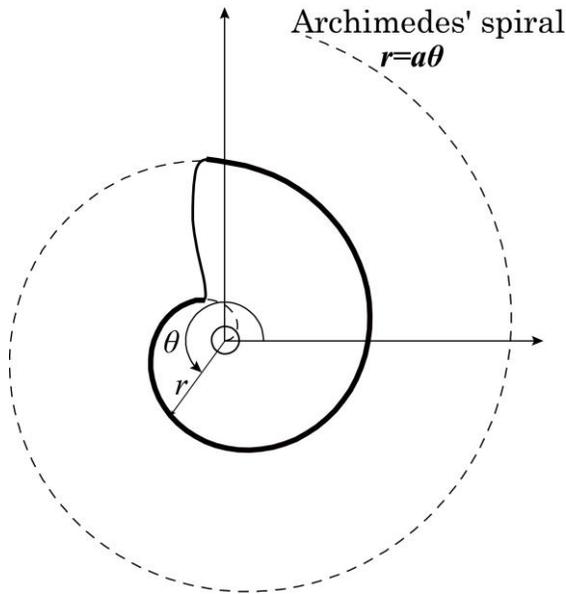

Fig. 7. Slope of non-circular pulley.

$$\begin{cases} F = kx & (3) \\ F = rmg/R & (4) \end{cases}$$

$$x = R\theta \quad (5)$$

$$r = a\theta \quad (6)$$

$$a = kR^2/mg \quad (7)$$

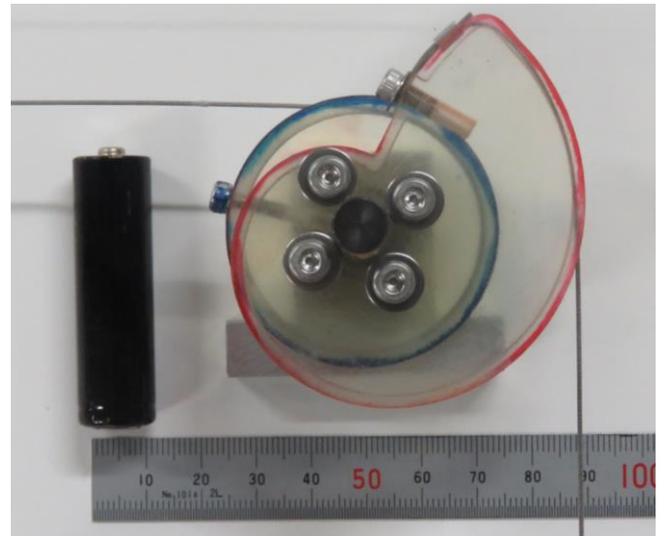

Fig. 10. The non-circular pulley.

4. ACTUAL MACHINE EXPERIMENT

4.1 Basic Performance Test

In order to confirm the effectiveness of the invented floating displacement / force converter, an experiment was carried out using an actual machine shown in Figure 11, 12 which was realized by combining a tension spring and a reverse spring. By inputting the displacement at the balance point and the force applied to the left end of the drawing spring in the figure as an output, we confirmed that the operation part can move right and left with a force smaller than the tensile force of the spring and the force change by the tension spring.

Next, using the actual machine, displacement and force of the operating part and generated force due to the spring were measured. A load cell was attached to the left side of the spring, and the operation force was measured using a force gauge constrained by a linear guide. Displacement of the operating part is the elongation when the natural length of the spring is 0. The experimental results are shown in Fig.13. It can be seen that when x = 0 mm, the spring generating force can be controlled with a small operating force of about 0.3%. In addition, when x = 10, 20 mm assuming application to a robot gripper, a certain force is required for the operating force, but it can be seen that the generated force can be controlled with the force of about 5% and 8%, respectively. Because it is a constant force, it is easier to control when using a motor etc, but it was found by actual machine experiment that it is better to design x to be smaller.

Since the radius of the non-circular pulley is in the range of 10 - 40 mm this time, the region of low tensile force can not be reproduced. In the future, those with radius 0 - will also be realized, and inverse characteristics will be reproduced in the entire region of the spring(Fig. 14).

The actual motion is shown in the attached movie with this paper for easy understanding.

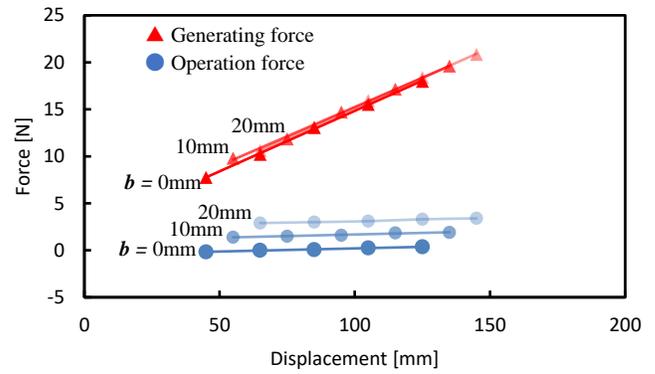

Fig. 13. Measured displacement and force.

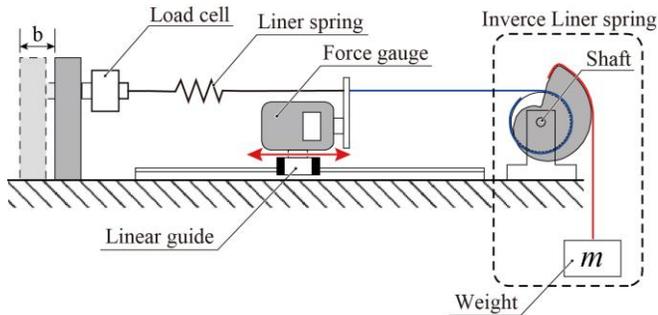

Fig. 11. Configuration diagram of experimental Device

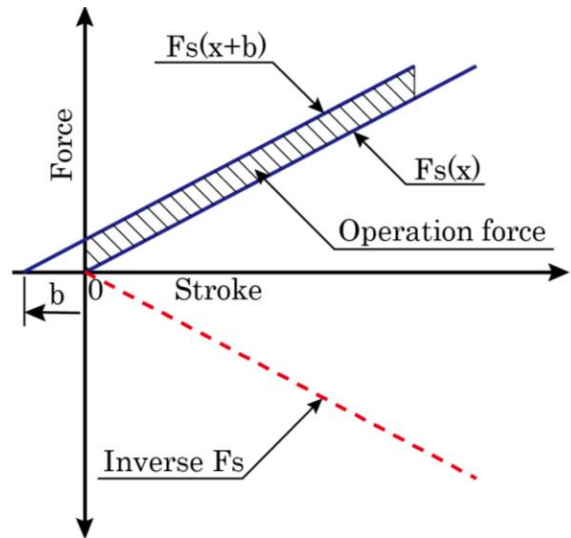

Fig. 14. Offset of the operational force(Fs) and inverse Fs

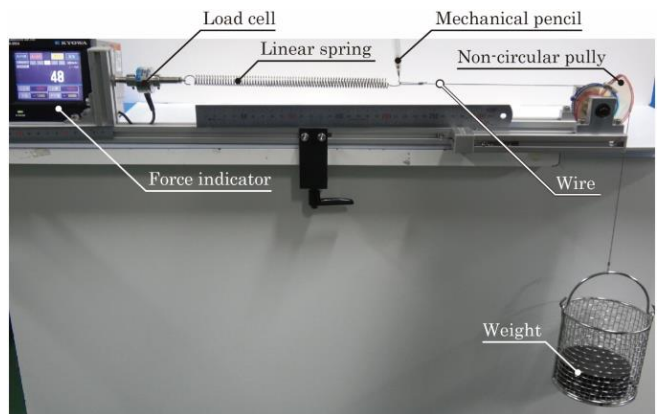

Fig. 12. Actual prototype model for the experiments.

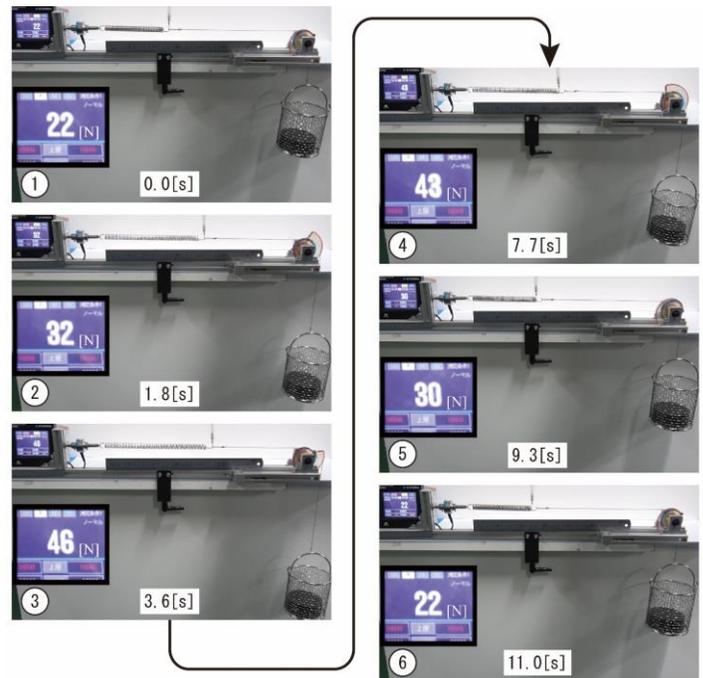

Fig. 15. Operation experiment.

## 4.2: Prototype Model with Small Actuator Type

The prototype model of the floating converter with small actuator for the purpose of the robotic mechanism as shown in Fig. 5 has been developed. The overview of the prototype model is shown in Fig. 16. The inner table shows the spec of the motor for the actuator of this prototype model. In addition, the experimental setup to see the performance of this mechanism is shown in Fig. 17. The experimental setup has the load cell to measure the force of the left end of the spring to show the force to grasped objects by this mechanism. The motion is shown in Fig. 18. Just only relatively small and weak actuator, large force can be realized by using this converter mechanism as shown in Fig. 18. The actual motion is shown in the attached movie with this paper for easy understanding.

Through the experiments with the actual prototype model the effectiveness of the proposed converter has been confirmed.

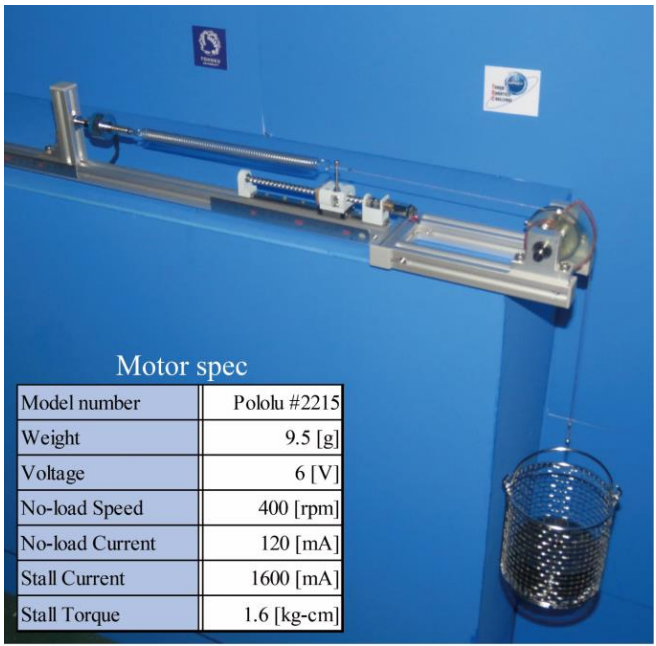

Fig. 16. Experimental Setup

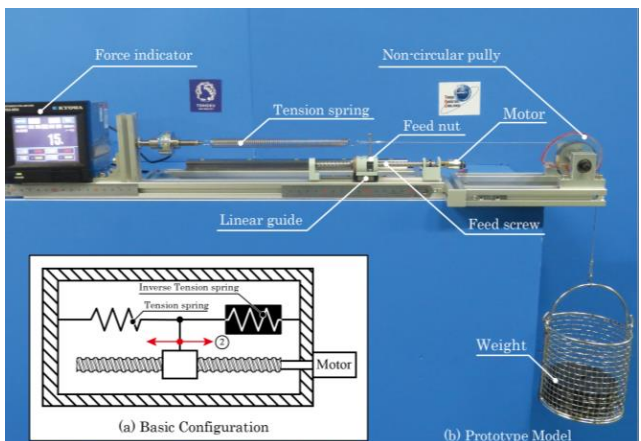

Fig. 17. Conversion Mechanism with built-in actuator

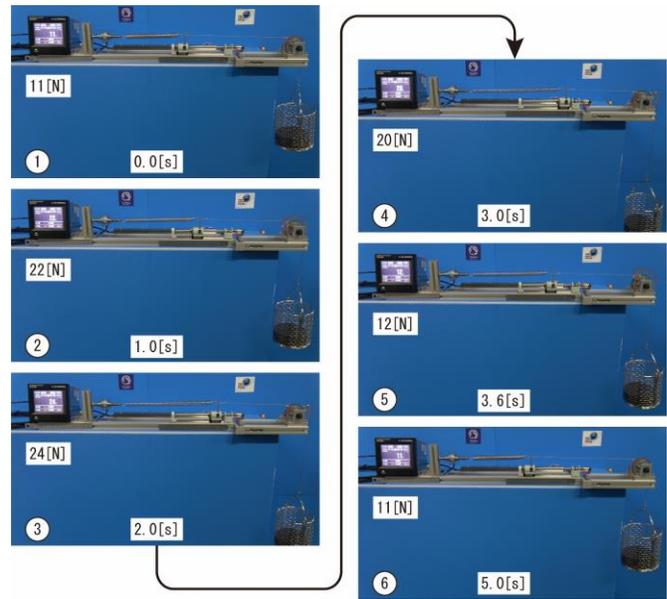

Fig. 18. Operation experiment with built-in actuator.

## 5. CONCLUSION

In this paper, we devised a mechanism capable of converting displacement and force, combining an elastic body having certain displacement and force characteristics and an elastic body having the opposite characteristic to manipulate the balance point. First, in order to confirm the effectiveness of the inventive principle, a prototype inverse characteristic spring was combined with a linear spring to verify the principle of a floating displacement / force converter. In addition, we measured the operating force and confirmed that we can control spring generating force with small force and displacement.

In the future, we will prototype a reverse spring using a lighter and more responsive spring.

As future prospects, the inventive mechanisms are: (1) a robot gripper mechanism that achieves high speed and high gripping force and has small energy consumption in force control, (2) a rigidity switching according to the tensile force of the internal wire with a linear body It can be expected to be applied to a one - dimensional jamming dislocation mechanism [9], 3) a brake mechanism, 4) a pressure regulating valve mechanism using a spring in an internal structure, and the like.